\documentclass{article} 
\usepackage{amsmath}  
\usepackage{multicol}  
\usepackage{graphicx}   
\usepackage{ragged2e}   
\usepackage{amsthm}
\usepackage{booktabs}
\usepackage{multirow}
\usepackage{array}
\usepackage{float}   
\usepackage{caption}     
\usepackage{subfigure}   
\usepackage{fancyhdr}
\usepackage{indentfirst} 
\usepackage{geometry}
\usepackage{cuted}
\usepackage{setspace}
\usepackage{listings}   
\usepackage{xcolor}
\usepackage{url}
\usepackage{chngcntr}
\counterwithin{figure}{section}
\counterwithin{table}{section}
\begin{document}  
\title{\LARGE\bf Deep Autoencoder Model Construction Based on Pytorch}	
\author{Junan Pan, Zhihao Zhao}
\date{}		
\maketitle     

\begin{spacing}{1.5}
\large{\bf Abstract:}    This paper proposes a deep autoencoder model based on Pytorch. This algorithm introduces the idea of Pytorch into the auto-encoder, and randomly clears the input weights connected to the hidden layer neurons with a certain probability, so as to achieve the effect of sparse network, which is similar to the starting point of the sparse auto-encoder. The new algorithm effectively solves the problem of possible overfitting of the model and improves the accuracy of image classification. Finally, the experiment is carried out, and the experimental results are compared with ELM, RELM, AE, SAE, DAE.
     

 \section{Introduction}
 
An image is a picture formed by the real existence of the outside world through the visual system of the brain. Since the image is the most intuitive thing that humans see and the easiest to understand, the image is the main means for people to obtain information from the outside world. The key work of image recognition is to explore the representation of the input image, and finally let the machine complete the understanding of the input image by itself, so as to recognize and classify it. Image processing processes externally input samples through machines, and filters out important features from the input images to realize image recognition and classification. Image recognition involves a wide range of research contents, such as license plate recognition, handwritten digit recognition, face recognition, recognition and classification of parts in machining, and accurate weather forecasting based on meteorological satellite photos. Among them, handwritten digit recognition mainly studies the digits handwritten by humans on paper, and uses machines to automatically complete the recognition and classification. Handwritten numbers are often encountered in our daily life, such as: manual bills, hand-filled express orders, and old-fashioned bank deposit slips, etc. It can be seen that the study of handwritten number classification has important practical application value.
 
 \section{Learning Algorithms of Neural Networks}
\subsection{Convolutional neural network} 
Convolutional neural network is a multi-layer neural network, which is composed of neurons with self-learning weights and biases, each neuron receives some input, and does some dot product calculations, and the output is each classification It is only a change in the function and form of the layer, it is still a neural network, and it is an improvement of the traditional neural network. Its artificial neurons can correspond to a part of the surrounding cells in the coverage area, and have excellent performance for large images. The default input of convolutional neural network CNN is an image, which can encode specific properties into the network structure, making the feedforward function more efficient and reducing a large number of parameters. Because the input of the convolutional neural network is a picture, the neurons are designed as three dimensions of width, height and depth (this depth is not the depth of the neural network, but the depth of the neuron). Convolutional neural network is mainly composed of convolution layer, excitation layer, pooling layer and fully connected layer. Their input and output are three-dimensional, as shown in Figure 2-1, which is the volume used by Yann LeCun in handwritten font recognition.
\begin{figure}[H]
  \centering
  \includegraphics[width=12cm]{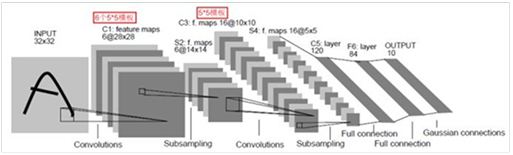}
  \caption{Convolutional Neural Network Structure}
\end{figure}

\subsubsection{Convolutional layer} 
Convolutional layer, each layer of the convolutional neural network consists of several convolutional units, and each convolutional unit is optimized by the back-propagation algorithm. The purpose of the convolution calculation is to extract different features of the input from each layer. The first layer of convolution may only extract some low-level, simple features such as lines, edges, and corners, and the subsequent layers can iteratively extract from the low-level features. complex features.
Once parameter sharing is applied, the computation on each layer is the convolution of the input layer and the weights, which is where the convolutional neural network gets its name. For example, an image with a size of 5×5 and a convolution kernel of 3×3, the convolution kernel here has 9 parameters, denoted as, in this case, the convolution kernel has 9 neurons, and the output forms a A 3×3 matrix, called a feature map. The first neuron is connected to the first 3×3 part of the image, and the second neuron is connected to the second part (the process will overlap). Specifically, as shown in Figure 2-2, the upper part of the figure is the output of the first neuron, and the lower part is the output of the second neuron. The calculation of each neuron is $f(x)=\operatorname{act}\left(\sum_{i j}^{n} \theta_{(n-i)(n-j)} x_{i j}+b\right)$.

\begin{figure}[H]
  \centering
  \includegraphics[width=8cm]{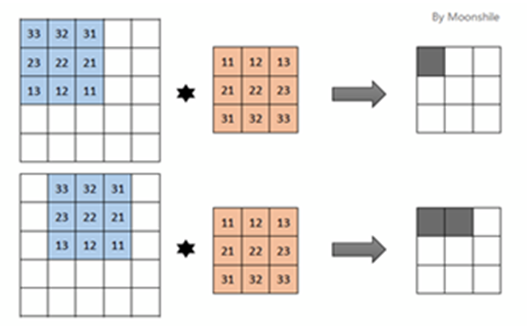}
  \caption{Convolution process}
\end{figure}

 \subsubsection{Incentive Layer}
The excitation layer (ReLUctant layer), which non-linearly maps the output of the convolutional layer. The current excitation functions generally include Sigmoid, Tanh, ReLU, Leaky ReLU, ELU, Maxout, etc. The commonly used excitation functions in convolutional neural networks are generally ReLu and its variant Leaky RelU. The commonly used excitation function is generally the Rectified Linear Unit (ReLU), because it has the advantages of fast convergence and simple gradient calculation. The ReLU function is shown in Figure 2-3.

\begin{figure}[H]
  \centering
  \includegraphics[width=7cm]{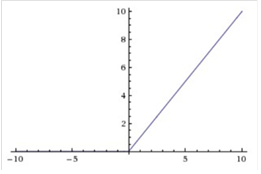}
  \caption{ReLU function}
\end{figure}

\subsubsection{Pooling layer}
The pooling layer is also called the downsampling layer. Pooling mainly reduces the amount of computation by aggregating features and reducing dimensionality. Generally speaking, features with large dimensions will be obtained after convolution, which requires a large amount of computation. In order to reduce the amount of computation, the pooling layer will cut the features into several regions, and take their maximum or average value to obtain features with smaller dimensions.

The pooling operation is independent for each depth slice, and the size is generally 2×2. The calculation methods of the pooling layer generally include four types: Max Pooling, Mean Pooling, Gaussian Pooling and Trainable Pooling. The commonly used methods are Max Pooling and Mean Pooling. Maximum pooling: take the maximum value of 4 points; mean pooling: take the average of 4 points.

The most common pooling layer is of scale 2×2 with stride 2, and down-sampling is performed on each depth slice of the input. Each Max operation is performed on four numbers, as shown in Figure 2-4. After the pooling operation, the depth is unchanged. If the size of the input unit of the pooling layer is not an integer multiple of 2, you can fill in the value to make up the multiple of 2, and then perform the downsampling operation.
\begin{figure}[H]
  \centering
  \includegraphics[width=8cm]{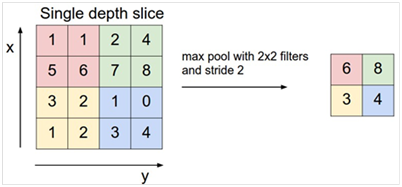}
  \caption{Pooling process}
\end{figure}

\subsubsection{Fully connected layer}
The fully-connected layer integrates all local features into global features. After the input signal undergoes multiple convolution, excitation and pooling operations, the output is multiple groups of signals. After the full connection operation, the multiple groups of signals are combined into one group of signals at a time. The global features obtained after going through such a layer by layer can be used to calculate the score of the last category.

The fully connected layer and the convolutional layer can be converted to each other. To turn the convolutional base layer into a fully connected layer only needs to turn the weights into a huge matrix, most of which are 0 except for some specific blocks (because of local perception), What's more, there are still many blocks with different weights. the same. For any fully connected layer, it can also become a convolutional layer, such as a fully connected layer with K=4096, the input layer size is 7×7×512, it can be equivalent to a F=7, Convolutional layer with P=0, S=1, K=4096F=7, P=0, S=1, K=4096. In other words, we set the filter size to be exactly the size of the entire input layer.

 \subsection{Deep Residual Network}
Deep convolutional neural networks have made major breakthroughs in image classification. They can naturally integrate low-level, intermediate-level, and high-level features. Generally speaking, the deeper the network, the richer the learned features. Recent evidence shows that the network The effect of depth on network performance is critical, and the main results on the challenging ImageNet dataset all employ very deep models, ranging from 16 to 30 layers deep. So does it mean that simply stacking deeper networks will perform better? However, the first obstacle to this problem is the infamous gradient vanishing and gradient exploding problems, which hinder the convergence of the network. Later, researchers found that this problem can be alleviated by normalizing the input data and batch normalization, which is generally not a problem for a dozen-layer network.

Therefore, in the paper, He et al. designed the following two simple stacked networks, a simple stacked 20-layer network and a simple stacked 50-layer network. Except for the different depths, the basic network blocks are the same, He et al. used cifar-10 as the training set. The experiments showed that for a simple stacked network, the deeper the depth, the higher the training error and the test error. This suggests that this phenomenon is not due to overfitting, as the deeper the network on the training set, the larger the error. That is to say, simply stacking the network to increase the depth of the network does not improve the performance of the network. He et al. call this phenomenon the degradation problem, which shows that not all systems are easy to optimize. For a shallow network, if the network achieves a certain performance, we superimpose a new network layer (referred to as the deep network) on the basis of the network, then the deep network should not be worse than the previous network, because if the shallow network already has the best performance If so, so many superimposed networks can learn that the latter are all identity mappings. But in fact, the training error is also rising, then it is the problem of the network itself, that is, simply stacking deeper networks makes the network itself difficult to train.

In response to this problem, the concept of deep residual learning came into being. In 2015, He et al. proposed the ResNet network. The network needs to be optimized for the residual module. Compared with the expectation that each stacked layer can directly learn the desired mapping function, the residual module explicitly allows the stacked layer to learn the residual. The mapping function, that is, the hidden mapping function expected to be learned is expressed as H(x),and the nonlinear mapping function expected to be learned by the network is set as F(x)=H(x)-x,  that is, the original expected learned The mapping function is rewritten as F(x)+x. The implementation of this structure is to add a channel (control feedforward) on the basis of the original network. Here, the newly added channel needs to skip one or more network layers. It can be seen that new network blocks do not need to add new ones. parameter. Experiments show that learning the residual map is easier than learning the original map. In extreme cases, if the identity map is optimal, it is much easier to learn all-zero values through stacked network layers than to learn an identity map.

 \section{Research on Deep Autoencoder Algorithm}
\subsection{Pytorch Theory}
Pytorch sparses the network structure dynamically, and the difference between them is that the sparseness is done differently. Figure 3-1 is a schematic diagram of Dropout. When training the Dropout method, the output value calculated by the activation function of the hidden layer node is randomly cleared during the forward transmission; Pytorch is the input connected to the hidden layer neurons. The weights are randomly cleared with a certain probability. Figure 3-2 is a schematic diagram of Pytorch. In other words, the fully connected layer becomes a sparsely connected layer using the Pytorch method, and its connections are randomly selected during the training phase. It should be noted that this does not mean that the weight matrix is set to a fixed sparse matrix during the training process, because during each training, the weights are randomly cleared, not fixed each time.
\begin{figure}[H]
  \centering
  \includegraphics[width=8cm]{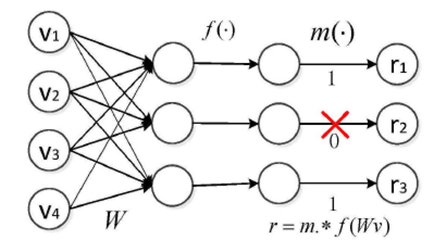}
  \caption{Dropout structure}
\end{figure}

\begin{figure}[H]
  \centering
  \includegraphics[width=8cm]{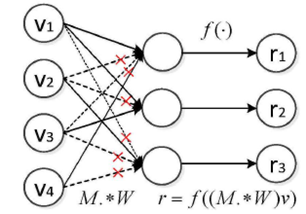}
  \caption{Pytorch structure}
\end{figure}

In the action layer of the Pytorch method, the output expression is:

\begin{equation}
r=a((M * W) v)
\end{equation}

Among them, v is the input sample matrix, W is the weight matrix between the input layer and the hidden layer, M is a column vector of 0, 1 values ($M_{ij}$-Bernoulli(P)), used to achieve random clearing of weights, f (•) is an activation function, and f(•) satisfies f(0)=0. During training, the elements of the pair matrix of each sample are independent, and essentially each sample corresponds to a different connection.
\begin{figure}[H]
  \centering
  \includegraphics[width=12cm]{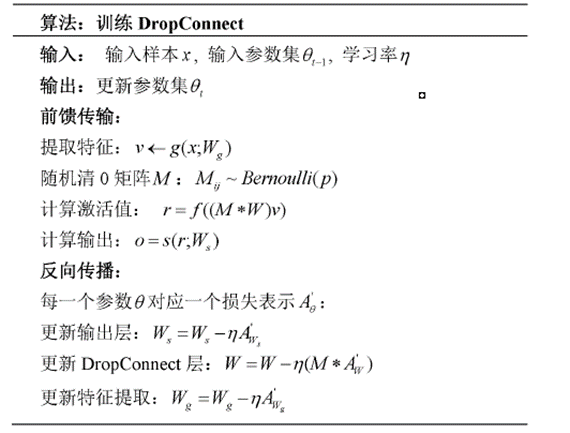}
  \caption{training process of Pytorch}
\end{figure}

\subsection{Pytorch-based deep autoencoder model}
\subsubsection{DDAE Model structure}
Autoencoders usually use a stacked structure in practical applications. The application of stacked autoencoders eliminates the huge workload of manually extracting data features, improves the efficiency of feature extraction, and exhibits powerful learning capabilities. Our model employs 2-layer AE, as shown in Figure 3-4. And the idea of Pytorch is introduced in the first layer of AE. Pytorch is a new regularization method proposed by Wan et al. on the basis of Dropout, which can effectively improve the test results of the algorithm. As can be seen from the figure, after the introduction of the Pytorch idea, the input layer and the hidden layer are no longer all connected, but randomly connected, so that the dynamic sparseness of the network can be achieved.

\begin{figure}[H]
  \centering
  \includegraphics[width=10cm]{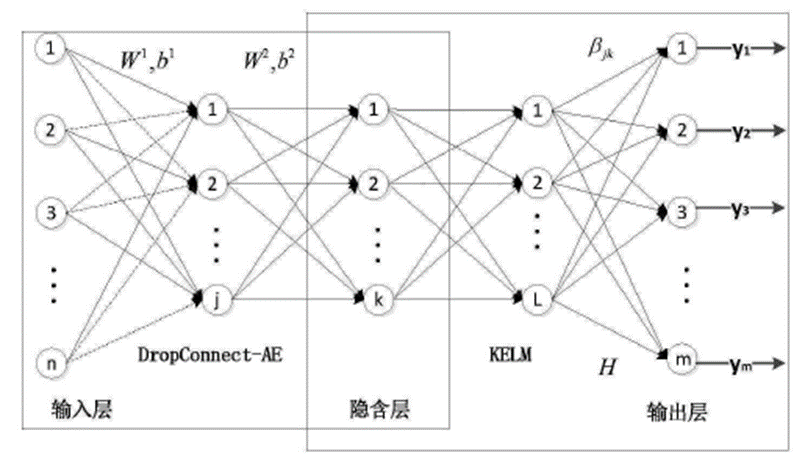}
  \caption{DDAE model structure}
\end{figure}

Since AE does not yet know how to connect input samples to classification, it cannot implement sample classification yet. In order for AE to implement the classification function, a classifier needs to be added to the top layer. Usually BP, softmax, SVM, etc. are used, but the training of these structures is very long, so we try to find a more convenient and effective classifier. The classifier chosen for our deep model here is KELM.

\subsubsection{Training process}
The deep model structure always contains five layers, which are trained by two AE superimpositions. Its specific training process is as follows:

(1) $x^{l}$ represents the images used for training, learn the weights and biases of the first hidden layer nodes using the Pytorch-Autoencoder method. Different from AE, the Pytorch regularization method is added to the hidden layer, that is, the input weights connected to the hidden layer are randomly cleared with a probability of 1-p.

\begin{equation}
M=\left(\text { rand }\left(\operatorname{size}\left(W_{1}\right)\right)>d r o p c o n n e c t F r a c t i o n\right)
\end{equation}

where M is the binary connection matrix $M_{ij}$-Bernoulli(p), $W_{1}$ is the weight matrix between the input layer and the first hidden layer, and pytorchFraction is a decimal from 0 to 1.

coding:
\begin{equation}
y^{1}=s\left(\left(M * W_{1}\right) x^{1}+b_{1}\right)
\end{equation}

decoding:
\begin{equation}
z^{1}=s\left(W_{1}^{T} y^{1}+b_{1}^{\prime}\right)
\end{equation}

The Pytorch-Autoencoder method requires $z^{l}$ as much as possible the same as the input $x^{l}$, and its loss function is shown in formula (5):
\begin{equation}
L\left(x^{1}, z^{1}\right)=\sum_{i=1}^{n} K L\left(x_{i}^{1} \| z_{i}^{1}\right)
\end{equation}

The difference between the input $\left(x_{1}^{b}, x_{2}^{b}, \ldots\right)$ and the output $\left(z_{1}^{b}, z_{2}^{b}, \ldots\right)$ is measured by the divergence value $K L\left(x_{i}^{l} \| z_{i}^{l}\right)$.

The stochastic gradient descent algorithm is used to train the weights of Pytorch-Autoencoder, and the weights are updated using formula (5).

\begin{equation}
W_{1}=W_{1}-n \frac{\partial L\left(x^{1}, z^{1}\right)}{\hat{\partial} W_{1}}
\end{equation}

The update method of the offset b, is the same as the formula (6);

After the above training, the parameters $W_{1}$  and $b_{1}$  have been trained, we can calculate the output features of the first hidden layer:
\begin{equation}
x^{2}=s\left(W_{1} x^{1}+b_{1}\right)
\end{equation}

(2) The second AE is trained below. The rough calculated above serves as the input value. The parameter training process is as follows:

coding:
\begin{equation}
y^{2}=s\left(W_{2} x^{2}+b_{2}\right)
\end{equation}
where and are the connection weights and biases between the first hidden layer and the second hidden layer.

decoding:
\begin{equation}
z^{2}=s\left(W_{2}^{T} y^{2}+b_{2}^{\prime}\right)
\end{equation}

AE requires $z^{2}$ as much as possible to be the same as the input $x^{2}$ , through the divergence value $K L\left(x_{i}^{2} \| z_{i}^{2}\right)$, to measure the difference between the input and the output.

The stochastic gradient descent algorithm is used to train the AE parameters $w_{2}$ and $b_{2}$ , and the parameter update formula is the same as formula (6). The values of $w_{2}$ and $b_{2}$ have been obtained through the above training, so the calculation formula of the output value of the third layer of the model is as follows:
\begin{equation}
x^{3}=s\left(W_{2} x^{2}+b_{2}\right)
\end{equation}

 (3) Finally, KELM is used for classification calculation, the input value is the output result of the third layer, and the classification training is carried out in a supervised learning method.
The calculation formula of the output matrix H of the fourth layer of DDAE is as follows:
\begin{equation}
H=\left[\begin{array}{c}
h\left(x_{1}^{3}\right) \\
\vdots \\
h\left(x_{N}^{3}\right)
\end{array}\right]_{N \times L}
\end{equation}

The variable L in the formula is the number of nodes in the fourth layer, and N is the number of input samples. Next, the output of the hidden layer of the sample $h\left(x_{i}^{3}\right)$, is regarded as a nonlinear mapping of the sample. When this mapping is unknown, we replace the kernel matrix $H H^{T}$, by constructing a kernel function. The Gaussian kernel function is a kernel function often used by researchers, so the Gaussian kernel function is selected in the experiment:
\begin{equation}
H H^{T}(i, j)=K\left(x_{i}^{3}, x_{j}^{3}\right)=\exp \left\{-\left\|x_{i}^{3}-x_{j}^{3}\right\|^{2} / 2 \gamma^{2}\right\}
\end{equation}

\begin{equation}
H H^{T}= \Omega _{ELM} =\left[\begin{array}{c}
0 \qquad\qquad\cdots \qquad\exp \left\{-\left\|x_{1}^{3}-x_{N}^{3}\right\|^{2} / 2 \gamma^{2}\right\}\\
\vdots \qquad  \qquad\ddots \qquad\qquad\qquad\vdots\\ 
\exp \left\{-\left\|x_{N}^{3}-x_{1}^{3}\right\|^{2} / 2 \gamma^{2}\right\}\qquad\qquad\cdots\qquad 0
\end{array}\right]
\end{equation}

\begin{equation}
h\left(x^{3}\right) H^{T}=\left[\begin{array}{c}
\exp \left\{-\left\|x^{3}-x_{1}^{3}\right\|^{2} / 2 \gamma^{2}\right\} \\
\vdots \\
\exp \left\{-\left\|x^{3}-x_{N}^{3}\right\|^{2} / 2 \gamma^{2}\right\}
\end{array}\right]
\end{equation}

where $\gamma$ is the kernel parameter, and it is necessary to select the optimal value through multiple experiments.

The weight matrix of the hidden layer and the output layer $\beta$ is expressed as formula (15), and the classification formula is expressed as formula (16).

\begin{equation}
\beta=\left(\frac{I}{C}+\Omega_{E M M}\right)^{-1} Y
\end{equation}

\begin{equation}
\left.f\left(x^{3}\right)=g\left(h\left(x^{3}\right) H^{T} \beta\right)=g\left(\left[\begin{array}{c}
K\left(x^{3}, x_{1}^{3}\right) \\
\vdots \\
K\left(x^{3}, x_{N}^{3}\right)
\end{array}\right]\left(\frac{I}{C}+\Omega_{E L M}\right)^{-1} Y\right)\right)
\end{equation}

\subsection{Experiment and result analysis}
In order to evaluate the performance of the deep model, we first conduct image reconstruction experiments on the MNIST dataset to verify the reconstruction ability of Pytorch-AE; then we conduct image classification experiments on the USPS and MNIST datasets respectively. Experimental environment: 64-bit operating system, 18 GB RAM, core 17 DMIZ-Intel X79PCH-mainboard, Intel Xeon ES-16200 3.6GHz processor, MATLAB 2012b.
\subsubsection{Image reconstruction}
In order to prove the reconstruction ability of the model after AE introduces the idea of Pytorch, we use the original AE and the deformed structure SAE and DAE to perform image reconstruction experiments with the algorithm Pytorch-AE in this paper. Validated on MNIST, all 60,000 training samples and 10,000 test samples were selected for the experiment; all images in the dataset were processed so that the size of each image was 28×28 pixels.
Figure 3-5 is the weight diagram of the hidden layer of each algorithm, and Figure 3-6 is the image reconstructed by AE, SAE, DAE and oropconnect.

\begin{figure}[H]
  \centering
  \includegraphics[width=10cm,height=10.50cm]{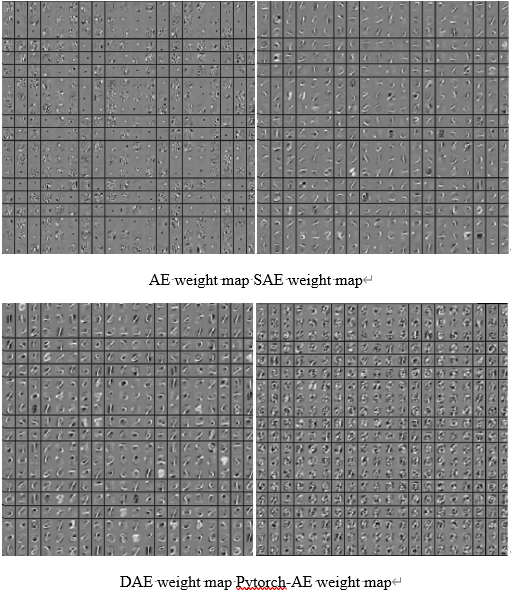}
  \caption{Hidden layer weight diagram}
\end{figure}

\begin{figure}[H]
  \centering
  \includegraphics[width=10cm,height=10.5cm]{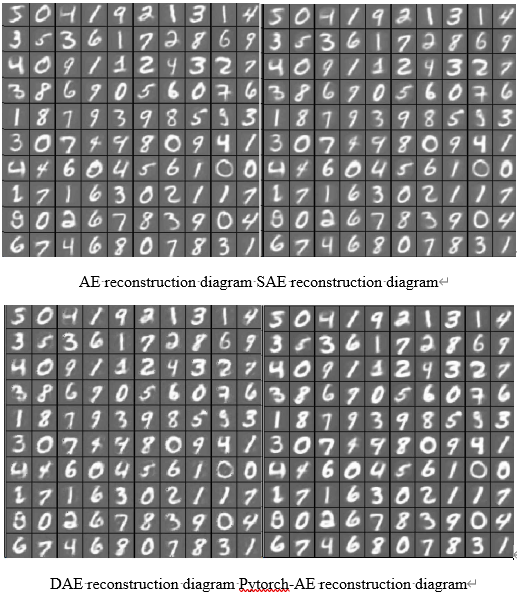}
  \caption{Reconstructed image}
\end{figure}

It can be seen from the figure that the Pytorch-AE model has good reconstruction ability, which can deepen the image texture and obtain the significant information of the image. It is proved that Pytorch-AE is feasible. In order to better prove its effectiveness, an image classification experiment will be carried out below.

\subsubsection{USPS dataset}
The USPS dataset is the U.S. Postal Service handwritten digit recognition library, with 7291 samples for training and 2007 samples for final testing. Here we select all training samples and test samples for experiments. The handwritten digit pictures in the dataset are all grayscale processed, so that the size of the images are all 16 and 16 pixel values.Table 3-1 is the accuracy comparison results in the USPS dataset.

\begin{table}[H]
\renewcommand\arraystretch{1.5}  
\centering
\caption{Accuracy comparison results in the USPS dataset}
\begin{tabular}{llcll}
\hline
Data & Algorithm            & \multicolumn{1}{l}{\#} & Training accuracy & Test accuracy \\
\hline
USPS & \multirow{2}{*}{ELM} & average                & 99.78(±0.03)      & 93.70(±0.32)  \\
     &                      & best                   & 99.85             & 94.57         \\
     & \multirow{2}{*}{RELM}& average                & 99.96(±0.004)     & 94.63(±0.27)  \\
     &                      & best                   & 99.97             & 95.22         \\
     & KELM                 & -                      & 99.97             & 95.42         \\
     & AE                   & -                      & 100               & 94.34         \\
     & SAE                  & -                      & 100               & 94.89         \\
     & DAE                  & -                      & 100               & 95.33         \\
     & DDAE                 & -                      & 100               & 96.31         \\
\hline
\end{tabular}
\end{table}

We use stacked AE, SAE, and DAE for image classification experiments, and compare the results with the model in this paper. Since the classifier of the deep model uses KELM, it is necessary to compare the experimental results with ELM and various variants of RELM and KELM, which is more convincing. Table 3-1 lists the experimental results comparison of ELM, RELM, KELM, AE, SAE, DAE and this model on the USPS dataset. As can be seen from Table 3-1, the test accuracy of the model in this paper is not only higher than that of the single hidden layer ELM, RELM, KELM, but also compared with the stacked AE, SAE, DAE depth models, the accuracy is also significantly improved, which proves that The model is feasible and valid.
\subsubsection{MNIST dataset}
Due to the limitations of the experimental environment, this paper randomly selects 6000 samples from the training samples for training, and randomly selects 1000 samples from the test samples for testing. As above, the original AE and some deformed SAE and DAE of AE are used for image classification experiments, and the results are compared with the model in this paper. Since KELM was selected for the classifier in the experiment, the same experiment was compared with ELM and its variants RELM and KELM. Table 3-2 lists the classification accuracy values. It can be seen from Table 3-2 that the model in this paper has achieved an accuracy of 97.64$\%$, which is obviously better than other models and achieves the results we expected, which further proves that the deep model DDAE is feasible and effective.

\begin{table}[H]
\renewcommand\arraystretch{1.5}  
\centering
\caption{Accuracy comparison of focus in MNIST data}
\begin{tabular}{llcll}
\hline
Data  & Algorithm            & \multicolumn{1}{l}{\#} & Training accuracy & Test accuracy \\
\hline
MNIST & \multirow{2}{*}{ELM} & average                & 99.92(±0.03)      & 92.27(±0.52)  \\
      &                      & best                   & 99.98             & 93.20         \\
      & \multirow{2}{*}{RELM}& average                & 99.82(±0.004)     & 94.50(±0.36)  \\
      &                      & best                   & 99.90             & 95.30         \\
      & KELM                 & -                      & 100               & 95.80         \\
      & AE                   & -                      & 100               & 95.30         \\
      & SAE                  & -                      & 100               & 95.60         \\
      & DAE                  & -                      & 100               & 95.73         \\
      & DDAE                 & -                      & 100               & 97.64         \\
\hline
\end{tabular}
\end{table}

 \section{Conclusion}
This paper proposes a hybrid autoencoder model DDAE, which is proposed in this paper for the classifier selection problem of deep learning. The idea of Pytorch solves the problem of possible overfitting of the model and improves the accuracy of image classification. Compared with AE, ELM, KELM and other algorithms, the DDAE model fully combines the effective feature extraction of Pytorch-based autoencoder and the fast and efficient characteristics of KELM. Experiments show that the model can effectively improve the classification accuracy.

\section*{References}
[1] Li Yanzhi, Chen Changhong, Xie Xiaofang. Research on aurora image classification algorithm based on improved convolutional neural network [J]. Journal of Nanjing University of Posts and Telecommunications (Natural Science Edition), 2019,39(06):86-93.

[2] Yang Mengzhuo, Guo Mengjie, Fang Liang. Research on Image Classification Algorithm Based on Keras Convolutional Neural Network [J]. Science and Technology Wind, 2019(23):117-118.

[3] Gao Lei, Fan Bingbing, Huang Sui. An Improved Convolutional Neural Network Image Classification Algorithm Based on Residuals [J]. Computer System Applications, 2019, 28(07): 139-144.

[4] Dong Yiming. Research on vehicle recognition algorithm based on distributed convolutional neural network [J]. Henan Science and Technology, 2019(20):28-31.

[5] Zhang Sianqing. Image classification algorithm based on convolutional neural network [J]. Computer Products and Circulation, 2019(06):112.

[6] Gao Zixiang, Zhang Baohua, Lv Xiaoqi, Gu Yu. Two-way Convolutional Neural Network Image Classification Algorithm Based on Adaptive Pooling [J]. Computer Engineering and Design, 2019, 40(05): 1334-1338.

[7] Qi Yongfeng, Li Fayong. Hyperspectral image classification algorithm based on locally preserved dimensionality reduction convolutional neural network [J]. Journal of Agricultural Machinery, 2019, 50(03): 136-143.

[8] Yang Zhenzhen, Kuang Nan, Fan Lu, Kang Bin. Overview of Image Classification Algorithms Based on Convolutional Neural Networks [J]. Signal Processing, 2018,34(12):1474-1489.

[9] Jiang Wenchao, Liu Haibo, Yang Yujie, Chen Jiafeng, Sun Aobing. A High Similarity Image Recognition and Classification Algorithm Integrating Wavelet Transform and Convolutional Neural Network [J]. Computer Engineering and Science, 2018, 40(09): 1646 -1652.

\end{spacing}   


\end{document}